\documentclass{article}

\usepackage{arxiv,times}

\usepackage{amsmath,amsfonts,bm}

\def\eqref#1{equation~\ref{#1}}

\def\1{\bm{1}}

\DeclareMathAlphabet{\mathsfit}{\encodingdefault}{\sfdefault}{m}{sl}
\SetMathAlphabet{\mathsfit}{bold}{\encodingdefault}{\sfdefault}{bx}{n}

\usepackage{amsmath}
\usepackage{amssymb}
\usepackage{array}
\usepackage{booktabs}
\usepackage{enumitem}
\usepackage{float}
\usepackage{graphicx}
\usepackage{makecell}
\usepackage{microtype}
\usepackage{pgfplots}
\usepackage{xspace}
\usepackage[hidelinks]{hyperref}
\usepackage{url}
\usepackage{algorithm} 
\usepackage{algpseudocode}
\usepgfplotslibrary{groupplots}
\pgfplotsset{compat=1.18}

\newcommand{\method}{InstaGen}
\newcolumntype{L}[1]{>{\raggedright\arraybackslash}p{#1}}

\title{InstaGen: Group-Shared Importance Sampling for Fast Video Generation}

\author{%
Kislay Aditya Oj\textsuperscript{1,*}\hspace{0.6em}
Kumar Krishna Agrawal\textsuperscript{2,*}\hspace{0.6em}
Sri Surya Varma Datla\textsuperscript{1}\hspace{0.6em}
Aditya Desai\textsuperscript{1} \\[2pt]
\textsuperscript{1}Computer Science and Engineering, Indian Institute of Technology Bombay, India \\
\textsuperscript{2}Electrical Engineering and Computer Sciences, University of California, Berkeley, USA
}

\iclrfinalcopy  %

\begin{document}

\maketitle
\begingroup
\makeatletter
\long\def\@makefntext#1{\noindent#1}
\makeatother
\footnotetext{%
\textsuperscript{*}Equal contribution. Correspondence:
\href{mailto:kislay@cse.iitb.ac.in}{\texttt{kislay@cse.iitb.ac.in}},
\href{mailto:apdesai@iitb.ac.in}{\texttt{apdesai@iitb.ac.in}}
}
\endgroup
\vspace{-8pt}

\begin{figure*}[h]
\centering
\includegraphics[width=1\linewidth]{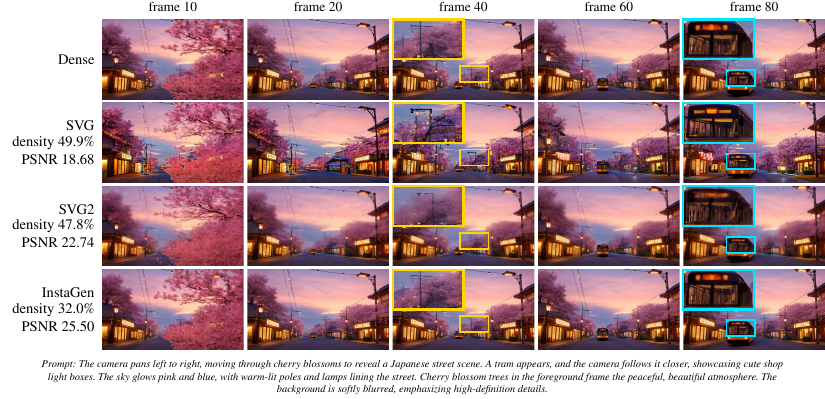}
\caption{\textbf{Frames from a single Wan2.1-T2V-1.3B generation (720p, 81 frames, 50 steps).} InstaGen achieves higher pixel fidelity, reaching a PSNR of 25.50 dB at a lower attention density of 32\% when compared to SVG and SVG2.}
\label{fig:teaser}
\end{figure*}

\begin{abstract}
\looseness=-1 Diffusion Transformers (DiTs) are the dominant architecture for video generation, but their quadratic attention complexity results in substantial computational cost and latency. Permuted block-sparse attention (PBSA) offers a promising approach to reducing this cost by selectively computing blocks of a semantically permuted attention matrix. However, existing approaches typically select only the top few key blocks for each query block, following a top-$p$ paradigm. We find that top-$p$ selection can be highly sample inefficient in high-entropy attention regimes, which are common in video generation. In these regimes, we show that sampling can provide substantially better approximations under the same computational budget. 
More generally, we cast block selection as importance sampling, where the top-scoring keys form an exact set that is computed in full and the remaining keys are represented by a sample drawn in proportion to their estimated attention and reweighted by inverse sampling probability.
Building on this formulation, we present InstaGen, which can be parameterized by any PBSA method: it replaces the method's block selection with an exact set plus a weighted sample. Each query cluster draws one sample shared by all its queries, so the sampled keys form blocks amenable to efficient GPU execution. 
In this paper, we take the SVG family of PBSA methods and show that, on Wan2.1-T2V-1.3B and Wan2.1-T2V-14B, InstaGen parameterized by SVG2 improves PSNR over SVG2 by 3.4 dB and 2 dB, respectively, at matched attention density, and by 1.6 dB and 2.2 dB at matched wall-clock time.
On Wan2.1-T2V-14B, InstaGen also beats SVG-EAR by 1.1 dB PSNR at nearly the same density while running faster (1.65× against 1.48× over dense).

\end{abstract}

\begin{figure}
\centering
  \includegraphics[width=0.8\textwidth]{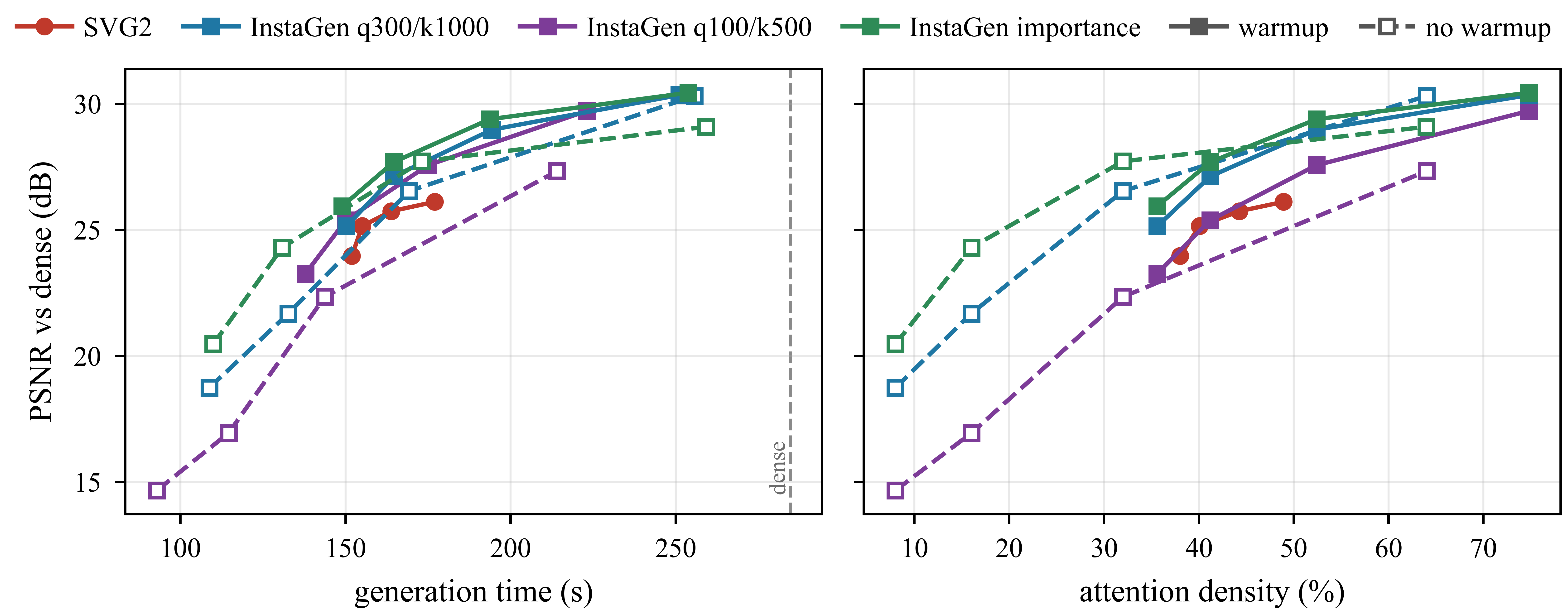}
  \caption{\textbf{\method{} gives higher quality than SVG2 at both matched density and matched latency on Wan2.1-T2V-1.3B.} \emph{Left:} PSNR against generation time per video. \emph{Right:} PSNR against attention density, the share of attention computed over the whole generation. SVG2 uses the configuration from its paper ($C_q{=}100$, $C_k{=}500$, full attention for the first 30\% of steps, key floor 0.1) at top-$p$ 0.5, 0.75, 0.9 and 0.95. InstaGen runs exact + sampled budgets from 4+4\% to 32+32\% at two clustering configurations for uniform and one for importance, with the same warmup and without it. 
   }
   \label{fig:frontier}
\end{figure}

\section{Introduction}
\label{sec:introduction}

\looseness=-1 Diffusion Transformers (DiTs) have been instrumental in achieving state-of-the-art quality in image and video generation \citep{wanteam2025wan, yang2025cogvideox, kong2024hunyuanvideo}. However, image and video generation with DiTs remains computationally expensive, primarily due to the quadratic cost of attention, which is repeated across layers and diffusion steps. As a result, generating even short videos can be extremely slow. For instance, generating a single 5-second 720p video (81 frames, 50 denoising steps) with Wan2.1-T2V-14B takes about 22.6 minutes on one H100 GPU. Improving generation latency is therefore essential for enabling the efficient generation of high-quality, long-form videos.

\looseness=-1 Training-free sparse attention is a natural paradigm for reducing the computational workload of attention in existing video generation models and improving generation latency. However, effective attention sparsity must strike a unique balance between three often non-aligned goals: (1) Quality per query: Each query token should, in effect, attend to the tokens that are most important for that specific query. (2) Efficiency: The sparsity pattern should be amenable to block-based execution so that it can naturally integrate with FlashAttention \citep{shah2024flashattention3} and benefit from hardware acceleration. (3) Alignment with the underlying model: While each query may, in principle, attend to its most important tokens, the approximation error introduced by sparsity can accumulate across layers and diffusion steps, potentially derailing the model and causing substantial downstream quality degradation. Permuted block-sparse attention (PBSA) methods such as SVG2 and SVG-EAR \citep{yang2025svg2,zhou2026svgear} have primarily focused on addressing the first two goals through clustering query and key tokens, permuting the tokens into contiguous blocks, scoring each block from the cluster centroids, and computing only the selected blocks with a block-sparse attention kernel. However, their treatment of the third goal often results in an unfavorable trade-off between sparsity and downstream generation quality.

\looseness=-1 
To maintain alignment, it is critical to ensure that the local approximation error in attention computation under sparsity is well controlled. Existing PBSA methods employ top-$p$ as the underlying sparsification strategy. However, for high-entropy attention distributions, top-$p$ ends up retaining a large number of tokens. Thus, the top-$p$ paradigm inevitably sacrifices either efficiency or approximation quality. In fact, to the best of our knowledge, all existing sparse video generation methods retain 30\% dense steps, as any sparsification under the top-$p$ paradigm leads to large approximation errors. This calls for a paradigm shift in sparsity. \method{} moves away from the deterministic sparsification underlying existing PBSA methods and instead employs stochastic sampling based on importance scores. We find that stochastic sparsity offers a significantly better quality-efficiency tradeoff. In fact, it enables us to completely eliminate dense steps, achieving fully sparse inference for video generation across all layers and all steps.

\looseness=-1 In \method{}, we formalize sparse attention as a sampling problem. For each query cluster, the keys are split into an exact set of top-scoring keys, whose attention is computed exactly, and a residual whose contribution is estimated from a sample of its keys reweighted by inverse sampling probability~\citep{horvitz1952generalization}. Top-k selection is the special case with no sample. We consider two ways of drawing this sample, importance and uniform sampling. In importance sampling, the attention estimated from the cluster centroids sets the sampling probabilities, and we draw the sample by proportional-to-size sampling without replacement (PPSWOR; Section~\ref{sec:ppswor}). Uniform sampling includes every residual key with equal probability, thus making it a special case of importance sampling. In this paper, we parameterize \method{} by SVG2, i.e., it keeps SVG2’s clustering and block scoring and replaces its top-$p$ selection. This choice is not a requirement, and the same estimator can be applied to other methods that produce block scores, such as SVOO~\citep{luo2026svoo}.

Sampling addresses the third goal, but if done naively it can conflict with the second, i.e., efficient execution. If each query had its own sample, it would attend to its own random keys. Then no two queries share a block of keys, and FlashAttention cannot run it. Drawing about 75,600 separate samples per head in every attention call (Wan2.1 at 720p, 81 frames) is also costly. We instead draw one sample per query cluster, using the probabilities computed from its centroid. All queries within the cluster share the resulting sampled keys, yielding a group-shared sampling scheme. This amortizes the sampling overhead across queries while producing a block structure: the cluster’s queries and the sampled keys form a dense block, which is compatible with FlashAttention and can be efficiently accelerated on GPUs. Moreover, sharing a sample adds no bias. For every query in the cluster, the estimates of the attention sums stay unbiased, because the weights use the probabilities the keys were actually drawn with.

We evaluate \method{} against SVG~\citep{xi2025sparsevideogen}, SVG2, and SVG-EAR on Wan2.1-T2V-1.3B and Wan2.1-T2V-14B, measuring fidelity to the dense model's video (PSNR, SSIM~\citep{wang2004ssim}, LPIPS~\citep{zhang2018lpips}) and video quality with VBench~\citep{huang2024vbench}. Over SVG2, \method{} improves PSNR by 3.4 dB and 2 dB on the 1.3B and 14B models, respectively, at matched attention density (Figure~\ref{fig:frontier}, Section~\ref{sec:experiments}), and on the 14B model it beats SVG-EAR by 1.1 dB PSNR at nearly the same density while running faster. On Wan2.1-T2V-1.3B, \method{} needs no dense warmup: it runs sparse attention at every diffusion step and still outperforms SVG and SVG2 run with their warmup (Figure~\ref{fig:teaser}). Removing the warmup matters most for long videos, because a dense step grows almost quadratically with video length: generating 561 frames, \method{} without warmup takes 57.3 minutes, against 82.8 for SVG2 and 162 for dense attention (Figure~\ref{fig:frames-quality}).

\section{Related Work} \label{sec:related}
\textbf{Structured sparse attention for video diffusion.}
Sparse VideoGen (SVG) profiles sampled query rows to assign each head a
spatial or temporal sparsity layout \citep{xi2025sparsevideogen}, and Radial
Attention uses a static $\mathcal{O}(S\log S)$ mask whose density decays
with temporal distance \citep{li2025radial}. Both fix the layout
structurally rather than adapting it to content.

\textbf{Content-adaptive sparse routing.}
Other methods estimate block importance from content: XAttention scores
blocks by antidiagonal sums \citep{xu2025xattention}, AdaSpa searches
hierarchical block patterns online \citep{xia2025training}, and
SpargeAttention filters blocks with compressed query/key representatives
and an in-kernel softmax-aware test \citep{zhang2025spargeattention}. SVG2
clusters queries and keys, permutes clusters into contiguous storage, and
selects cluster pairs by a centroid score \citep{yang2025svg2}; AdaCluster
adapts the clustering to layerwise token geometry
\citep{tan2026adacluster}, and SVOO combines an offline per-layer sparsity
schedule with online query--key co-clustering \citep{luo2026svoo}. These
methods route each query cluster by a single centroid-derived score.
\method{} reuses this component to score token importance.

\textbf{Recovering skipped attention.}
SVG-EAR approximates skipped blocks with key/value centroids and recomputes
those with high predicted error \citep{zhou2026svgear}; SparsePR learns an
affine residual correction from the computed attention
\citep{taghavi2026sparsepr}; and Sol-Attn estimates the skipped contribution
from the computed subset \citep{li2026solattn}. vAttention, developed for
LLM decoding, instead samples the residual, giving an unbiased estimate
with a probabilistic error guarantee \citep{desai2026vattention}. We extend
this idea by casting sparse attention as importance sampling, under which
top-$k$ selection and residual sampling arise naturally.

\textbf{Changing the model.}
Veda distills full-attention tile distributions into a learned selector
\citep{han2026veda}; SANA-Video trains a block-linear video diffusion
transformer \citep{chen2025sanavideo}, and SANA-Video 2.0 a hybrid with
periodic softmax anchors and cross-depth attention residuals
\citep{chen2026sanavideo2}. We focus on training-free sparse attention.

\section{Background}
\label{sec:background}

\subsection{Scaled Dot-Product Attention}
\label{sec:sdpa}

Consider a single self-attention head with queries, keys, and values
$Q,K,V\in\mathbb{R}^{n\times d}$, where $n$ is the sequence length and $d$ is
the head dimension. For query $i$, the scaled attention logits are
\begin{equation}
  \ell_{ij} = \frac{q_i^\top k_j}{\sqrt{d}}, \qquad j\in\{1,\dots,n\}.
\end{equation}
Softmax attention then produces the output
\begin{equation}
  o_i
  = \sum_{j=1}^{n} \alpha_{ij}\, v_j,
  \qquad
  \alpha_{ij}
  = \frac{e^{\ell_{ij}}}{\sum_{j'=1}^{n} e^{\ell_{ij'}}},
\end{equation}
or, equivalently, as the ratio of two sums
\begin{equation}
  o_i = \frac{n_i}{z_i},
  \qquad
  n_i = \sum_{j=1}^{n} e^{\ell_{ij}}\, v_j,
  \qquad
  z_i = \sum_{j=1}^{n} e^{\ell_{ij}}.
  \label{eq:num-den}
\end{equation}
Evaluating dense attention for all $n$ queries costs $O(n^2 d)$ per head.
Sparse and sampled approximations target the sums in
Eq.~\eqref{eq:num-den}: rather than including every key/value index $j$, they
retain a small subset of indices and estimate the contribution of the
remainder. The sampling estimators described below apply directly to
approximating $n_i$ and $z_i$.

\subsection{Sampling with Probabilities Proportional to Size}
\label{sec:pps}

Consider the problem of estimating a sum $Y=\sum_{j=1}^{n}y_j$ from a subset of
$m < n$ of its terms. This problem underlies sparse attention, since the
numerator and denominator in Eq.~\eqref{eq:num-den} are sums of $n$ terms:
setting $y_j=e^{\ell_{ij}}v_j$ recovers $n_i$, and setting $y_j=e^{\ell_{ij}}$
recovers $z_i$. We assume that each term is associated with a score $s_j>0$ that
is much cheaper to compute than $y_j$ and serves as an estimate of its magnitude
$|y_j|$. The goal is an unbiased estimator of $Y$ with low variance under a
budget of $m$ terms.

\subsubsection{PPSWOR: Probabilities Proportional to Size Without Replacement}
\label{sec:ppswor}

PPSWOR is an importance sampling scheme that draws a sample $\mathcal{S}$ of $m$
distinct terms such that terms with larger scores, and hence a larger expected
contribution to the sum, are more likely to be included. The scheme is
characterized by the inclusion probabilities $\pi_j=\Pr[j\in\mathcal{S}]$, which
satisfy $\sum_{j=1}^{n}\pi_j=m$ and are set proportional to the scores,
$\pi_j\propto s_j$. Capping the target $m\,s_j/\sum_{j'}s_{j'}$ at $1$ and
renormalizing gives
\begin{equation}
  \pi_j=\min\big(1,\lambda s_j\big),
  \qquad
  \text{where } \sum_{j=1}^{n}\min\big(1,\lambda s_j\big)=m.
  \label{eq:pi}
\end{equation}
The left-hand side is nondecreasing in $\lambda$, so $\lambda$ is well defined.
The terms with $\pi_j=1$ form the heavy set
$\mathcal{H}=\{j:\lambda s_j\ge1\}$, which is the top-$k$ set by score, with
$k=|\mathcal{H}|$ determined by the scores rather than fixed in advance. The
sample therefore consists of all of $\mathcal{H}$ and $m-|\mathcal{H}|$ residual
terms, each included with probability exactly $\pi_j<1$.

The Horvitz--Thompson estimator~\citep{horvitz1952generalization} weights each
sampled term by the inverse of its inclusion probability. For the sample above,
it separates into a deterministic part and a sampled part:
\begin{equation}
  \hat{Y}
  = \underbrace{\sum_{j\in\mathcal{H}} y_j}_{\text{deterministic top-}k}
  + \underbrace{\sum_{j\in\mathcal{S}\setminus\mathcal{H}} \frac{y_j}{\pi_j}}_{\text{sampled residual}}.
  \label{eq:ht}
\end{equation}
This estimator is unbiased for any choice of scores, and top-$k$ selection is
recovered by dropping the second term. We use this formulation in \method{}.

\section{\method}

\label{sec:method}

\begin{algorithm}[t]
\caption{\method{} attention for one head. $\ell_{ij}=q_i^{\top}k_j/\sqrt{d}$ are the
logits of Section~\ref{sec:sdpa}, $\bar q_g$ and $\bar k_c$ are the centroids of query group $g$
and key cluster $c$, $c(j)$ is the cluster of key $j$, and $|c|$ is its size.
Budgets are fractions of the sequence length: $\rho_E{+}\rho_S$ means $m_E=\rho_E\,n$ and
$m_S=\rho_S\,n$ (written $E{+}S$, in percent, in Section~\ref{sec:experiments}). Warmup calls,
when used, compute dense attention instead.}
\label{alg:instagen}
\begin{algorithmic}[1]
\renewcommand{\algorithmicrequire}{\textbf{Input:}}
\renewcommand{\algorithmicensure}{\textbf{Output:}}
\Require $Q,K,V\in\mathbb{R}^{n\times d}$; budgets $m_E,m_S$; sampler \textsc{Uniform} or \textsc{Importance}
\Ensure $\hat{O}\in\mathbb{R}^{n\times d}$
\State Cluster queries into $C_q$ groups and keys into $C_k$ clusters
  \Comment{$k$-means, warm-started}
\State $s_{gc}\gets\exp\!\big(\bar{q}_g^{\top}\bar{k}_c/\sqrt{d}\big)$ for every pair $(g,c)$
  \Comment{centroid scores}
\For{each query group $g$}
  \State $\mathcal{E}_g\gets$ whole clusters by decreasing $|c|\,s_{gc}$ until
    $|\mathcal{E}_g|\approx m_E$ \Comment{exact keys}
  \State $\mathcal{R}_g\gets\{1,\dots,n\}\setminus\mathcal{E}_g$
    \Comment{residual keys}
  \If{sampler is \textsc{Uniform}}
    \State $\pi_j\gets m_S/|\mathcal{R}_g|$ for all $j\in\mathcal{R}_g$
    \State $\mathcal{S}_g\gets m_S$ keys of $\mathcal{R}_g$, uniformly without replacement
  \Else
    \State $\pi_j\gets\min\!\big(1,\lambda_g\,s_{g,c(j)}\big)$, with $\lambda_g$ such that
      $\sum_{j\in\mathcal{R}_g}\pi_j=m_S$ \Comment{\S\ref{sec:ppswor}}
    \State $\mathcal{S}_g\gets$ systematic PPS sample of $\mathcal{R}_g$ with probabilities $\pi_j$
  \EndIf
  \State $w_j\gets1$ for $j\in\mathcal{E}_g$, \ $w_j\gets1/\pi_j$ for $j\in\mathcal{S}_g$
    \Comment{Horvitz--Thompson}
  \State $\hat{o}_i\gets\sum_{j\in\mathcal{E}_g\cup\mathcal{S}_g}
    \operatorname{softmax}_j\!\big(\ell_{ij}+\log w_j\big)\,v_j$ for all $i\in g$
    \Comment{$=\hat{n}_i/\hat{z}_i$}
\EndFor
\State \Return $\hat{O}$
\end{algorithmic}
\end{algorithm}

\method{} builds on the observation in vAttention~\citep{desai2026vattention} that sampling can outperform top-$p$ selection in high-entropy regimes. \method{} formulates sparse attention as a sampling problem, which leads naturally to an importance sampling formulation with uniform sampling as a special case.

\method{} performs PPSWOR sampling using scores that serve as proxies for the exponentiated raw attention scores. We obtain these scores from clusters for two reasons: the centroid of a cluster is a cheap proxy that represents the scores of its tokens, and clustering lets us permute the tokens into contiguous blocks, so that attention follows the block structure of FlashAttention~\citep{shah2024flashattention3} and realizes the theoretical speedup of sparsity. Such scores can come from any method that produces them, such as SVG2~\citep{yang2025svg2} and SVOO~\citep{luo2026svoo}, among others. For clarity, we adopt the clustering of SVG2, in which the queries and keys are clustered separately with $k$-means into $C_q$ query groups and $C_k$ key clusters with centroids $\bar{q}_g$ and $\bar{k}_c$. The score of query group $g$ and key cluster $c$ is
\begin{equation*}
s_{gc} = \exp\!\big(\bar{q}_g^{\top}\bar{k}_c/\sqrt{d}\big),
\end{equation*}
which is the proxy for the exponentiated attention score of every query in $g$ with every key in $c$.

\begin{figure*}[t]
\centering
\includegraphics[width=\linewidth]{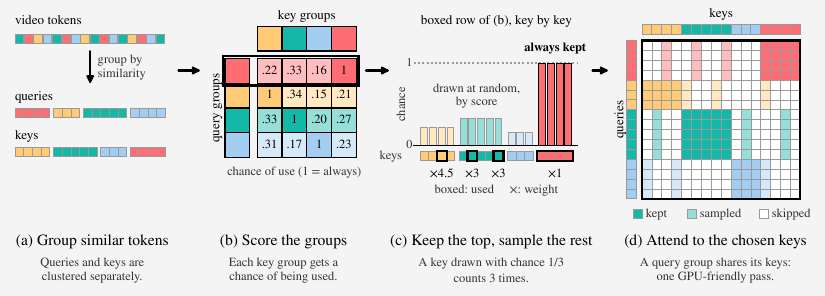}
\caption{\textbf{Overview of \method{}.} (a) Queries and keys are clustered into groups. (b) Centroid scores give each key group a probability $\pi$ of being used by each query group. (c) The top key group is always kept; the remaining keys are sampled in proportion to $\pi$ and weighted by $1/\pi$. (d) Queries in a group share the same keys, so attention runs over dense blocks in one kernel.}
\label{fig:instagen-attn-map}
\end{figure*}

\looseness=-1These scores are used to perform PPSWOR as described in Section~\ref{sec:ppswor}. All queries in a group share the same scores, and so do all keys in a cluster, so the whole procedure is carried out once per query group. For each query group, the total budget is split in advance into an exact budget $m_E$ and a sampling budget $m_S$. The exact budget is handled first: the key clusters with the largest total score are selected deterministically until $m_E$ is reached, and we set $\pi_j=1$ for their keys.

PPSWOR with budget $m_S$ is then applied to the remaining keys, the residual. Each residual key $j$ receives an inclusion probability $\pi_j$ proportional to its score and capped at 1, as in \eqref{eq:pi}, such that the sample contains $m_S$ keys. Keys whose probability reaches 1 are promoted, that is, they are always retained like the exact keys; since scores are constant within a cluster, whole clusters are promoted. The remaining keys ($\pi_j<1$) are sampled using PPS, and the sample is shared across all queries of the group. In our kernel, it is drawn with systematic PPS sampling, which attains the prescribed inclusion probabilities exactly. The resulting attention map is illustrated in Figure~\ref{fig:instagen-attn-map}.

This sharing across queries is what keeps the kernel GPU-friendly: all queries in a group attend to the same keys, so queries and attended keys form dense blocks. Let $\mathcal{S}$ be the PPS sample over the residual. For a query $i$, the estimators of the numerator $n_i$ and denominator $z_i$ in \eqref{eq:num-den} can be written as
\begin{equation*}
\hat{z}_i = \sum_{j=1}^{n} \left(
\mathbb{I}(\pi_j = 1)\, e^{\ell_{ij}}
+
\mathbb{I}(\pi_j < 1)\, \mathbb{I}(j \in \mathcal{S})\,
\frac{1}{\pi_j}\, e^{\ell_{ij}}
\right)
\end{equation*}
\begin{equation*}
\hat{n}_i = \sum_{j=1}^{n} \left(
\mathbb{I}(\pi_j = 1)\, e^{\ell_{ij}} v_j
+
\mathbb{I}(\pi_j < 1)\, \mathbb{I}(j \in \mathcal{S})\,
\frac{1}{\pi_j}\, e^{\ell_{ij}} v_j
\right),
\end{equation*}
where $\mathbb{I}(\cdot)$ is the indicator function, and the output is $\hat{o}_i=\hat{n}_i/\hat{z}_i$. Note that both the numerator and denominator estimators are unbiased:
\begin{equation*}
\mathbb{E}(\hat{z}_i) = z_i
\qquad
\mathbb{E}(\hat{n}_i) = n_i.
\end{equation*}

\paragraph{Hardware Efficiency.} \method{} is designed to be compatible with FlashAttention~\citep{shah2024flashattention3}. All queries in the same cluster attend to the same set of keys, whether exact or sampled. Thus, attention is computed in blocks rather than per query (each selected 128-key tile is either entirely exact or entirely sampled), with the importance weights applied as a logit bias. The sample is drawn on the GPU, where we implement the attention kernel by modifying FlashAttention-3 for NVIDIA Hopper GPUs. Clustering, scoring, and sampling run as separate GPU steps before the attention kernel. Figure~\ref{fig:instagen-attn-map}(d) illustrates this implementation.

\paragraph{Residual Approximation.} Like SVG-EAR~\citep{zhou2026svgear}, \method{} estimates the attention of skipped keys instead of dropping it. SVG-EAR does so deterministically, by replacing each skipped cluster with its centroid key and mean value. \method{} instead samples actual skipped keys with inverse-probability weights, giving unbiased estimates of the attention sums whose variance decreases as the sample grows.

\paragraph{Discussion on Alternative Sampling Schemes.} Several sampling schemes can be used to estimate the residual. While we use PPSWOR in our kernel, other choices are possible, and we discuss them below.

\textbf{PPSWR.} Sampling with replacement can be used to estimate the residual in Algorithm~\ref{alg:instagen}. Its downside is that keys may be sampled repeatedly. While this matters little statistically at small budgets, it becomes undesirable at large ones; for instance, it does not recover dense attention even when the budget equals that of dense attention. On the other hand, it is easier to analyze: the independence of the samples allows us to apply strong tail bounds, which can be used to devise a \textit{verifiable} approximation of attention for sparse video generation~\citep{desai2026vattention}.

\textbf{PPSWOR.} Sampling without replacement has several desirable properties for our algorithm. For instance, at large budgets, PPSWOR is guaranteed to cover a larger number of distinct keys. In the limit, it covers all keys for each query and recovers exact dense attention. However, its downside is that the variance of the estimator is difficult to analyze and, consequently, so are analytical guarantees on the approximation error.

\textbf{Poisson sampling.} Poisson sampling is another technique for residual sampling, in which each key is included independently by a separate Bernoulli trial. Its advantage is that the sample is obtained without replacement; however, the sample size equals the budget only in expectation. Nevertheless, the independence again allows us to apply strong tail bounds to the estimators, and thus to obtain a \textit{verifiable} sparse attention approximation.

These alternative schemes can be seen as extensions of this work, and we leave their exploration to future work.

\section{Experiments}
\label{sec:experiments}
\textbf{Models.} We evaluate \method{} on two open-source text-to-video
diffusion transformers, Wan2.1-T2V-1.3B and Wan2.1-T2V-14B~\citep{wanteam2025wan},
generating 81-frame videos at 720p ($720\times1280$). After 3D-VAE
tokenization, each video has 21 latent frames of 3{,}600 tokens, so every
self-attention call spans 75{,}600 tokens. We sparsify only self-attention;
cross-attention to the text embeddings remains dense.

\begin{table}[!t]
  \centering
  \caption{\textbf{Main results on Wan2.1-T2V-1.3B and 14B.} PSNR, SSIM and LPIPS compare each video with the dense video from the same seed, and all quality metrics are averaged over the 22 VBench prompts. $E$+$S$: exact and sampled share of keys per sparse call with dense warmup, as prescribed by each baseline's paper; \method{} uses SVG2's (first 30\% of steps and 3\% of layers). Density is overall, counting warmup calls as full attention. Speedup is end-to-end wall-clock on one H100 against a dense run. Bold: best sparse method in each column.}
  \label{tab:main}
  \vspace{4pt}
  \small
  \setlength{\tabcolsep}{4pt}
  \resizebox{\textwidth}{!}{%
  \begin{tabular}{lcccccccccccc}
    \toprule
    & & & \multicolumn{3}{c}{Dense-reference quality} & \multicolumn{4}{c}{VBench} & \multicolumn{2}{c}{Efficiency} \\
    \cmidrule(lr){4-6}\cmidrule(lr){7-10}\cmidrule(lr){11-12}
    Method & $E$+$S$ & Warm-up & PSNR$\uparrow$ & SSIM$\uparrow$ & LPIPS$\downarrow$ &
    ImgQual$\uparrow$ & AesQual$\uparrow$ & SubCons$\uparrow$ & BackCons$\uparrow$ &
    Density$\downarrow$ & Speedup$\uparrow$ \\
    \midrule
    \multicolumn{12}{l}{\textit{Wan2.1-T2V-1.3B}} \\
    Dense & -- & & -- & -- & -- & 0.6689 & 0.6416 & 0.9569 & 0.9612 & 100\% & 1.00$\times$ \\
    SVG & -- & \checkmark & 23.62 & 0.794 & 0.214 & 0.6837 & 0.6378 & 0.9489 & 0.9518 & 49.85\% & 1.47$\times$ \\
    SVG2 ($p{=}0.95$) & -- & \checkmark & 26.11 & 0.834 & 0.122 & 0.6185 & 0.6091 & 0.9420 & 0.9495 & 48.92\% & 1.65$\times$ \\
    SVG-EAR & -- & \checkmark & 28.29 & \textbf{0.910} & 0.064 & 0.6625 & 0.6357 & 0.9497 & 0.9543 & 34.87\% & 1.63$\times$ \\
    \addlinespace[2pt]
    \method{} (uniform) & 8+8 & \checkmark & 27.12 & 0.856 & 0.093 & 0.6331 & 0.5798 & 0.9411 & 0.9467 & 41.20\% & \textbf{1.76}$\times$ \\
    \method{} (uniform) & 16+16 & & 26.54 & 0.876 & 0.088 & 0.6718 & \textbf{0.6490} & 0.9521 & 0.9596 & \textbf{32.00\%} & 1.68$\times$ \\
    \method{} (uniform) & 16+16 & \checkmark & 28.98 & 0.906 & 0.057 & 0.6434 & 0.5976 & 0.9528 & 0.9572 & 52.40\% & 1.49$\times$ \\
    \addlinespace[2pt]
    \method{} (importance) & 8+8 & \checkmark & 27.69 & 0.869 & 0.071 & 0.6826 & 0.6272 & 0.9665 & 0.9591 & 41.20\% & 1.72$\times$ \\
    \method{} (importance) & 16+16 & & 27.72 & 0.891 & 0.061 & \textbf{0.6909} & 0.6312 & \textbf{0.9702} & \textbf{0.9673} & \textbf{32.00\%} & 1.63$\times$ \\
    \method{} (importance) & 16+16 & \checkmark & \textbf{29.39} & 0.906 & \textbf{0.053} & 0.6864 & 0.6306 & 0.9701 & 0.9652 & 52.40\% & 1.46$\times$ \\
    \midrule
    \multicolumn{12}{l}{\textit{Wan2.1-T2V-14B}} \\
    Dense & -- & & -- & -- & -- & 0.6876 & 0.6676 & 0.9677 & 0.9715 & 100\% & 1.00$\times$ \\
    SVG & -- & \checkmark & 21.01 & 0.742 & 0.202 & 0.6892 & \textbf{0.6742} & 0.9653 & 0.9728 & 51.11\% & 1.41$\times$ \\
    SVG2 ($p{=}0.9$) & -- & \checkmark & 22.38 & 0.749 & 0.167 & 0.6927 & 0.6239 & 0.9661 & 0.9704 & 53.34\% & 1.53$\times$ \\
    SVG-EAR & -- & \checkmark & 23.55 & 0.827 & 0.126 & 0.6848 & 0.6700 & 0.9666 & 0.9677 & 42.19\% & 1.48$\times$ \\
    \addlinespace[2pt]
    \method{} (uniform) & 8+8 & \checkmark & 22.91 & 0.771 & 0.136 & 0.6963 & 0.6345 & \textbf{0.9763} & 0.9693 & 42.67\% & \textbf{1.67}$\times$ \\
    \method{} (uniform) & 16+16 & & 19.10 & 0.664 & 0.237 & 0.6792 & 0.6334 & 0.9736 & 0.9748 & \textbf{32.00\%} & 1.63$\times$ \\
    \method{} (uniform) & 16+16 & \checkmark & 23.55 & 0.796 & 0.122 & \textbf{0.6990} & 0.6361 & \textbf{0.9763} & 0.9711 & 53.59\% & 1.45$\times$ \\
    \addlinespace[2pt]
        \method{} (importance) & 8+8 & \checkmark & 24.61 & 0.871 & 0.083 & 0.6877 & 0.6701 & 0.9701 & 0.9701 & 42.67\% & 1.65$\times$ \\
    \method{} (importance) & 16+16 & & 21.84 & 0.811 & 0.138 & 0.6858 & 0.6623 & 0.9730 & \textbf{0.9753} & \textbf{32.00\%} & 1.58$\times$ \\
    \method{} (importance) & 16+16 & \checkmark & \textbf{25.50} & \textbf{0.887} & \textbf{0.072} & 0.6857 & 0.6652 & 0.9699 & 0.9716 & 53.59\% & 1.42$\times$ \\
    \bottomrule
  \end{tabular}}
\end{table}

\textbf{Metrics.} We measure how closely each sparse method reproduces the dense model's video with Peak Signal-to-Noise Ratio (PSNR), Structural Similarity Index Measure (SSIM)~\citep{wang2004ssim} and Learned Perceptual Image Patch Similarity (LPIPS)~\citep{zhang2018lpips}. Each method is compared with the dense video from the same pipeline on the same hardware. We assess perceptual quality with four VBench~\citep{huang2024vbench} dimensions: Imaging Quality (ImgQual), Aesthetic Quality (AesQual), Subject Consistency (SubCons) and Background Consistency (BackCons). Efficiency is measured by \emph{density}, the fraction of query-key pairs computed relative to full attention. Table~\ref{tab:main} reports overall density, counting dense warm-up calls as full attention. End-to-end efficiency is measured by wall-clock speedup over dense generation.

\begin{figure}[!b]
\centering
\includegraphics[width=1\linewidth]{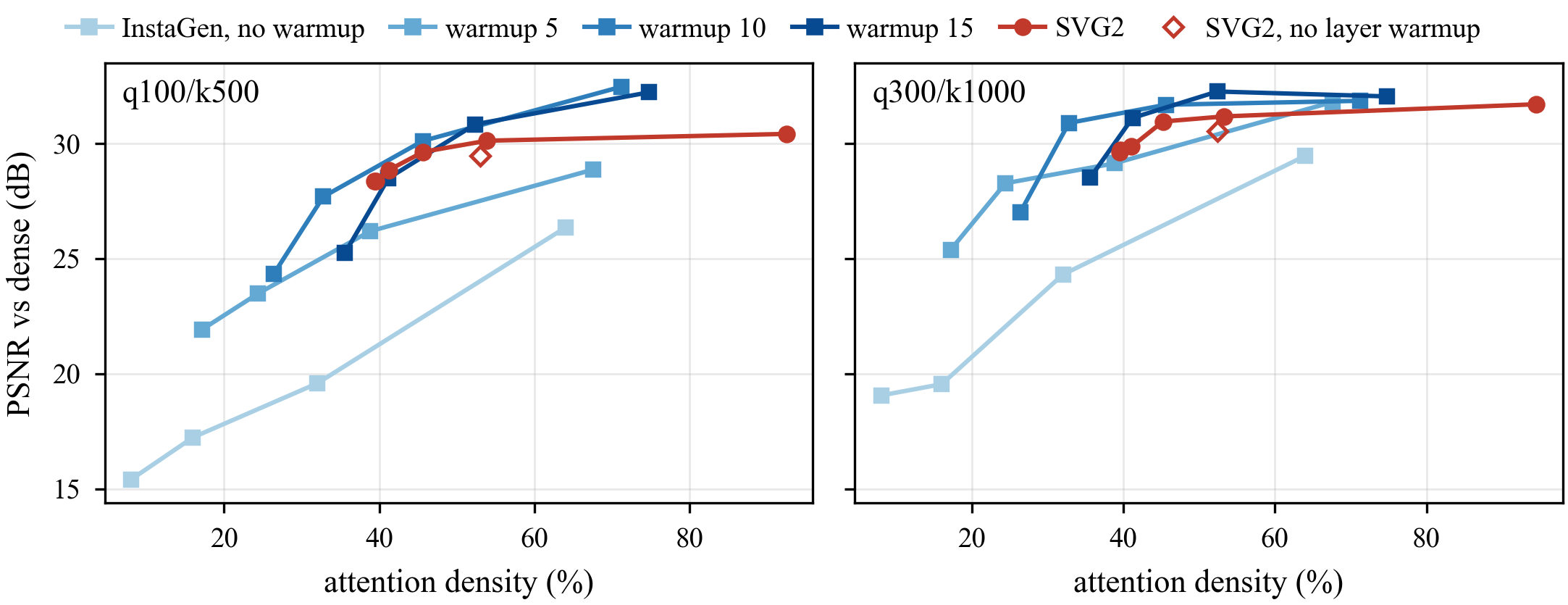}
\caption{ \textbf{Warmup ablations on Wan2.1-T2V-14B (720p, 81 frames,
50 steps).} \method{} with uniform sampling runs its first 0, 5, 10 or 15 of 50 steps at full attention
(warmup), ranging from 4+4\% to 32+32\% budget. SVG2 runs at its paper configurations at top-$p$ 0.1, 0.25, 0.5, 0.75, 0.9 and 0.99.  A 5--10 step warmup matches or beats the 15 steps SVG2 uses. The hollow diamond is SVG2 without its one-layer warmup; the \method{} runs have none. The prompt used is \texttt{svg2\_ex1}, Appendix~\ref{app:prompts}.} 
\label{fig:floor-ablation}
\end{figure}

\textbf{Prompts.} We use 22 prompts from the VBench prompt
suite~\citep{huang2024vbench}, the two prompts with the most motion from each of its eleven per-dimension lists, as judged by an LLM. Each prompt is generated once with a fixed seed, shared by all methods. Figures~\ref{fig:floor-ablation} and~\ref{fig:frames-quality}(b) use a single prompt, the first example prompt of SVG2's released Wan2.1 script~\citep{yang2025svg2}. Appendix~\ref{app:prompts} lists all prompts.

\textbf{Baselines.} We compare \method{} against dense attention and three
training-free sparse attention methods: the static method
SVG~\citep{xi2025sparsevideogen}, which assigns each head a spatial or temporal
pattern, and the dynamic methods SVG2~\citep{yang2025svg2}, which keeps the
$k$-means clusters that reach a top-$p$ attention-mass threshold, and
SVG-EAR~\citep{zhou2026svgear}, which approximates skipped blocks with centroids and routes by predicted error. Each baseline uses its official configuration and
dense warm-up schedule. We evaluate \method{} with both uniform and importance samplers to compare against these baselines.

\begin{figure}[t]
  \centering
  \includegraphics[width=\linewidth]{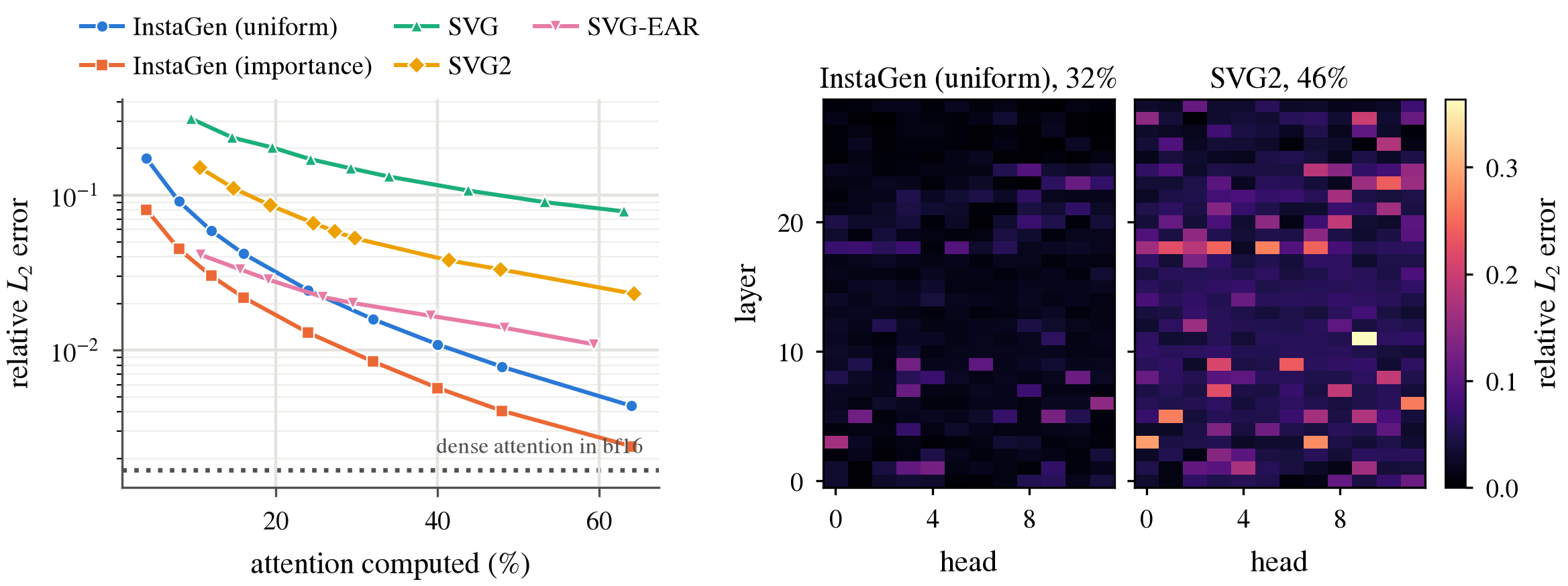}
      \caption{\textbf{\method{} approximates attention more accurately than prior
  sparse methods.}
  \emph{Left:} relative $L_2$ error of one self-attention call from
  Wan2.1-T2V-1.3B against exact attention. Importance sampling gives the lowest error at every density,
  several times below SVG2.
  \emph{Right:} per-block, per-head error over one full generation averaged across all steps.
  \method{} (uniform, 16+16, no warm-up) keeps the error low everywhere at 32\%
  density, while SVG2 ($p{=}0.9$, warm-up counted as zero error) leaves large
  errors in individual heads and layers despite computing 46\%.}
  \label{fig:error-pair}
\end{figure}

\textbf{Implementation.} We implement \method{} with a fused attention kernel
derived from FlashAttention-3 and benchmark on an NVIDIA H100 80GB SXM GPU. All
methods use 50 UniPC steps with flow shift 5.0 and guidance scale 5.0. We set
$C_q=300$ and $C_k=1000$, running 50 $k$-means iterations at a layer's first call
and two warm-started iterations thereafter~\citep{yang2025svg2}. Each query group attends exactly to
the top-scoring key blocks within $E\%$ of the keys and estimates the remainder
from $S\%$ of the keys, sampled in proportion to their estimated attention mass
(Section~\ref{sec:ppswor}); we denote this $E$+$S$. Both terms are computed in
one pass, with importance weights applied as a logit bias. Because \method{} estimates the omitted attention
rather than discarding it, it does not require the dense warm-up used by prior
methods. We report it both with and without warm-up, with the first
30\% of steps and 3\% of layers dense, as in SVG2.

\subsection{Quality and efficiency evaluation}
\label{sec:exp-quality}

\paragraph{Attention approximation.} We first experiment on one self-attention call captured from a Wan2.1-T2V-1.3B generation. Figure~\ref{fig:error-pair}(left) shows the mean relative error across all heads in one layer for different methods. Importance sampling has the lowest error at every density: at about 24\% of attention computed, its relative $L_2$ error is 0.013, against 0.066 for SVG2 and 0.022 for SVG-EAR, and at about 48\% it is 0.004, against 0.033 and 0.014. SVG is a static method, so the error comes via incorrect assumption of attention maps in different attention heads, while the error in other methods is the residual approximation of unselected keys. SVG2 fully drops them, so the error is larger; SVG-EAR and \method{} both apply different approximation techniques, which lead to relatively lower errors. Figure~\ref{fig:error-pair}(right) shows the error map across all the layers and heads for the whole generation. Averaged over all steps, \method{} (uniform, 16+16, no warmup) keeps every head and layer at low error at 32\% density, while SVG2 ($p{=}0.9$) leaves large errors in individual heads and layers despite computing 46\% even with warmup steps.

\begin{figure*}[!t]
\centering
\includegraphics[width=\linewidth]{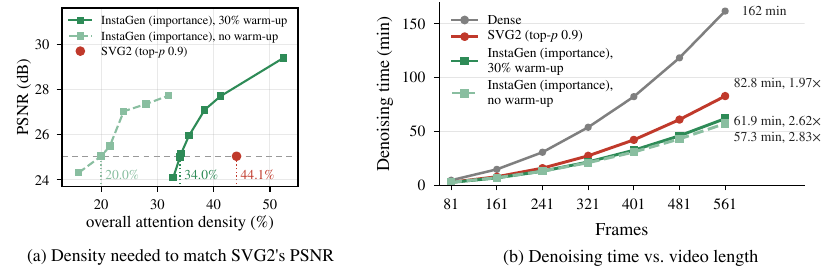}
\caption{\textbf{\method{} with importance sampling matches SVG2's quality at much lower density.} (a) PSNR against overall attention density (warm-up steps count as dense), averaged over 22 VBench prompts on Wan2.1-T2V-1.3B (720p, 81 frames, 50 steps). SVG2 in its paper configuration ($p{=}0.9$) reaches 25.03\,dB at 44.1\%; \method{} (importance sampling) reaches the same PSNR at 34.0\% with the same 30\% warm-up, and at 20.0\% with no warm-up. (b) Denoising time against the number of frames on one H100 SXM, on one prompt (\texttt{svg2\_ex1}, Appendix~\ref{app:prompts}; VAE decoding excluded), with \method{} (importance sampling) at 4+4\% with the 30\% warm-up and at 14+14\% with no warm-up, budgets at which it beats SVG2 ($p{=}0.9$) on PSNR, SSIM and LPIPS on this prompt at 81 frames. At 561 frames dense takes 162\,min, SVG2 82.8\,min (1.97$\times$), and \method{} 61.9\,min (2.62$\times$) with warm-up and 57.3\,min (2.83$\times$) without.}
\label{fig:frames-quality}
\end{figure*}

\paragraph{Video quality.} Table~\ref{tab:main} reports quality and efficiency across the two models. Both \method{} variants share the exact set and budget and differ only in how the sample is drawn: \emph{uniform} draws the shared sample uniformly without replacement, while \emph{importance} draws in proportion to the scores from the cluster centroids following PPSWOR. On both models, \method{} (importance, 8+8 with warm-up) improves PSNR over SVG2 by 1.6~dB at 1.3B and 2.2~dB at 14B while using lower density and running faster. On the 14B model it is the best sparse method on all three fidelity metrics, ahead of SVG-EAR at nearly the same density. Without any warm-up, \method{} still outperforms SVG and SVG2 on the 1.3B model, even though both run with warm-up. Importance sampling improves on uniform sampling at every configuration: drawing residual keys in proportion to their attention weight lowers the estimator's variance at the same budget.

 We compare algorithms at similar density to get a fair comparison of attention approximation of every method. We fix the overall density at that of SVG2's paper configuration ($p{=}0.9$) and give \method{} (importance) the same budget and warmup, split equally between exact and sampled keys. On Wan2.1-T2V-1.3B (44.1\%), \method{} raises PSNR from 25.03 to 28.42\,dB and SSIM from 0.780 to 0.884, and lowers LPIPS from 0.180 to 0.060; on Wan2.1-T2V-14B (54.0\%), from 23.40 to 25.41\,dB, 0.805 to 0.886, and 0.152 to 0.073. Figure~\ref{fig:frontier}(right) shows the curve for quality, where \method{} envelops SVG2 with both uniform and importance sampling.

\paragraph{Latency.}
Figure~\ref{fig:frontier} (left) plots quality against generation time on Wan2.1-T2V-1.3B. With importance sampling and no warmup, \method{} (16+16) reaches 27.72\,dB in 173.1\,s, 1.6\,dB above SVG2's best setting ($p{=}0.95$, 26.11\,dB in 177.0\,s); with warmup, \method{} (8+8) reaches 27.69\,dB in 164.6\,s, 1.6\,dB higher than SVG2's best in 12.4\,s less time. Figure~\ref{fig:frames-quality}(a) shows the density required to match the quality of SVG2, and at that matched quality Figure~\ref{fig:frames-quality}(b) shows the advantage of \method{} on longer frames, where it gets higher speedups than SVG2.

\paragraph{Warmup.} Sparse video methods keep their initial steps dense because early errors steer the whole trajectory. SVG2 runs its first 30\% of steps at full attention; SVG and SVG-EAR use similar schedules. Without any warmup, Figure~\ref{fig:frontier} shows that \method{} with importance sampling lies above SVG2's entire frontier, and on Wan2.1-T2V-14B (Figure~\ref{fig:floor-ablation}) it needs a short warmup, but 10 steps put it above SVG2 at every density. Removing the warmup matters more as videos get longer, because a dense step grows almost quadratically with length (4.6\,s at 81 frames, 187.6\,s at 561). In Figure~\ref{fig:frames-quality}(b), \method{} without warmup (14+14) is slower than with warmup (4+4) at 81 frames (164.2 vs.\ 153.2\,s), level at 241 (778.0 vs.\ 779.5\,s) and faster beyond, taking 57.3 against 61.9\,min at 561 frames.

\section{Conclusion}
\label{sec:conclusion}

This paper proposes \method{}, which replaces the top-$p$ block selection of permuted block-sparse attention with a sampling-based estimate of the full attention. At its core, \method{} computes an exact set of top-scoring keys in full and estimates the contribution of the remaining keys from a weighted sample, which keeps the estimates of the attention sums unbiased. Parameterized by SVG2, it reuses SVG2's clustering and block scores and replaces its top-$p$ selection. Group-shared sampling makes it fast: each query cluster draws one sample shared by all its queries, so the sampled keys form dense blocks that run with the exact blocks in a single FlashAttention pass.

Across sparse video generation experiments, \method{} consistently achieves a better quality-efficiency tradeoff than existing methods. On Wan2.1-T2V-1.3B and 14B, \method{} improves PSNR over SVG2 by 3.4\,dB and 2\,dB at matched attention density and by 1.6\,dB and 2.2\,dB at matched wall-clock time, and on 14B it beats SVG-EAR by 1.1\,dB at nearly the same density. On the 1.3B model it needs no dense warmup to outperform SVG2, which matters most for long videos: it denoises a 561-frame video in 57.3 minutes, against 82.8 for SVG2. Because \method{} treats sparse attention as sampling, other block scorers, such as SVOO's, can be plugged in. These results demonstrate that importance sampling provides an effective mechanism for allocating a limited attention budget between high-importance tokens and the residual attention mass, and offers a general framework that can be combined with different mechanisms for constructing proxy attention scores.

\bibliographystyle{iclr2027_conference}
\bibliography{references}

\begin{thebibliography}{23}
\providecommand{\natexlab}[1]{#1}
\providecommand{\url}[1]{\texttt{#1}}
\expandafter\ifx\csname urlstyle\endcsname\relax
  \providecommand{\doi}[1]{doi: #1}\else
  \providecommand{\doi}{doi: \begingroup \urlstyle{rm}\Url}\fi

\bibitem[Chen et~al.(2026{\natexlab{a}})Chen, Yu, Li, Xue, Liu, Xin, Zhao, Ye, Wu, Wang, Zhou, Luo, Han, and Xie]{chen2026sanavideo2}
Junsong Chen, Jincheng Yu, Yitong Li, Shuchen Xue, Haozhe Liu, Jingyu Xin, Yuyang Zhao, Tian Ye, Zhangjie Wu, Zian Wang, Daquan Zhou, Ping Luo, Song Han, and Enze Xie.
\newblock {SANA-Video 2.0}: Hybrid linear attention with attention residuals for efficient video generation, 2026{\natexlab{a}}.
\newblock URL \url{https://arxiv.org/abs/2607.21553}.

\bibitem[Chen et~al.(2026{\natexlab{b}})Chen, Zhao, Yu, Chu, Chen, Yang, Wang, Pan, Zhou, Ling, Liu, Yi, Zhang, Li, Chen, Cai, Fidler, Luo, Han, and Xie]{chen2025sanavideo}
Junsong Chen, Yuyang Zhao, Jincheng Yu, Ruihang Chu, Junyu Chen, Shuai Yang, Xianbang Wang, Yicheng Pan, Daquan Zhou, Huan Ling, Haozhe Liu, Hongwei Yi, Hao Zhang, Muyang Li, Yukang Chen, Han Cai, Sanja Fidler, Ping Luo, Song Han, and Enze Xie.
\newblock {SANA-Video}: Efficient video generation with block linear diffusion transformer.
\newblock In \emph{International Conference on Learning Representations}, 2026{\natexlab{b}}.
\newblock URL \url{https://arxiv.org/abs/2509.24695}.
\newblock Oral.

\bibitem[Desai et~al.(2026)Desai, Agrawal, Yang, Cuadron~Lafuente, Gaspar~Schroeder, Zaharia, Gonzalez, and Stoica]{desai2026vattention}
Aditya Desai, Kumar~Krishna Agrawal, Shuo Yang, Alejandro Cuadron~Lafuente, Luis Gaspar~Schroeder, Matei Zaharia, Joseph~E. Gonzalez, and Ion Stoica.
\newblock {vAttention}: Verified sparse attention via sampling.
\newblock In \emph{International Conference on Learning Representations}, pp.\  52253--52284, 2026.
\newblock arXiv:2510.05688.

\bibitem[Han et~al.(2026)Han, Yang, Mei, Hu, Jiang, and Qi]{han2026veda}
Shihao Han, Hao Yang, Xiaofeng Mei, Xinting Hu, Yi~Jiang, and Xiaojuan Qi.
\newblock {Veda}: Scalable video diffusion via distilled sparse attention.
\newblock In \emph{International Conference on Machine Learning}, 2026.

\bibitem[Horvitz \& Thompson(1952)Horvitz and Thompson]{horvitz1952generalization}
D.~G. Horvitz and D.~J. Thompson.
\newblock A generalization of sampling without replacement from a finite universe.
\newblock \emph{Journal of the American Statistical Association}, 47\penalty0 (260):\penalty0 663--685, 1952.

\bibitem[Huang et~al.(2024)Huang, He, Yu, Zhang, Si, Jiang, Zhang, Wu, Jin, Chanpaisit, Wang, Chen, Wang, Lin, Qiao, and Liu]{huang2024vbench}
Ziqi Huang, Yinan He, Jiashuo Yu, Fan Zhang, Chenyang Si, Yuming Jiang, Yuanhan Zhang, Tianxing Wu, Qingyang Jin, Nattapol Chanpaisit, Yaohui Wang, Xinyuan Chen, Limin Wang, Dahua Lin, Yu~Qiao, and Ziwei Liu.
\newblock {VBench}: Comprehensive benchmark suite for video generative models.
\newblock In \emph{Proceedings of the IEEE/CVF Conference on Computer Vision and Pattern Recognition}, pp.\  21807--21818, 2024.

\bibitem[Kong et~al.(2024)Kong, Tian, Zhang, Min, Dai, Zhou, Xiong, Li, Wu, Zhang, et~al.]{kong2024hunyuanvideo}
Weijie Kong, Qi~Tian, Zijian Zhang, Rox Min, Zuozhuo Dai, Jin Zhou, Jiangfeng Xiong, Xin Li, Bo~Wu, Jianwei Zhang, et~al.
\newblock {HunyuanVideo}: A systematic framework for large video generative models.
\newblock \emph{arXiv preprint arXiv:2412.03603}, 2024.

\bibitem[Li et~al.(2026)Li, Li, Chen, Ye, Liu, Yu, Wang, Zhang, Xie, Xie, and Han]{li2026solattn}
Haopeng Li, Yitong Li, Junsong Chen, Tian Ye, Haozhe Liu, Jincheng Yu, Duomin Wang, Ruihua Zhang, Zeke Xie, Enze Xie, and Song Han.
\newblock {Sol-Attn}: Accelerating video generation inference via on-the-fly attention sparsification, 2026.
\newblock URL \url{https://arxiv.org/abs/2607.24027}.
\newblock Technical report.

\bibitem[Li et~al.(2025)Li, Li, Cai, Xi, Yang, Lin, Zhang, Yang, Hu, Peng, Agrawala, Stoica, Keutzer, and Han]{li2025radial}
Xingyang Li, Muyang Li, Tianle Cai, Haocheng Xi, Shuo Yang, Yujun Lin, Lvmin Zhang, Songlin Yang, Jinbo Hu, Kelly Peng, Maneesh Agrawala, Ion Stoica, Kurt Keutzer, and Song Han.
\newblock Radial attention: $\mathcal{O}(n\log n)$ sparse attention with energy decay for long video generation.
\newblock \emph{arXiv preprint arXiv:2506.19852}, 2025.

\bibitem[Luo et~al.(2026)Luo, Chen, Wang, Wang, Zhu, Sun, Gao, Chen, and Li]{luo2026svoo}
Jiayi Luo, Jiayu Chen, Jiankun Wang, Cong Wang, Hanxin Zhu, Qingyun Sun, Chen Gao, Zhibo Chen, and Jianxin Li.
\newblock Attention sparsity is input-stable: Training-free sparse attention for video generation via offline sparsity profiling and online {QK} co-clustering, 2026.
\newblock URL \url{https://arxiv.org/abs/2603.18636}.
\newblock Accepted to ICML 2026.

\bibitem[Shah et~al.(2024)Shah, Bikshandi, Zhang, Thakkar, Ramani, and Dao]{shah2024flashattention3}
Jay Shah, Ganesh Bikshandi, Ying Zhang, Vijay Thakkar, Pradeep Ramani, and Tri Dao.
\newblock {FlashAttention-3}: Fast and accurate attention with asynchrony and low-precision.
\newblock In \emph{Advances in Neural Information Processing Systems}, volume~37, 2024.
\newblock \doi{10.52202/079017-2193}.

\bibitem[Taghavi et~al.(2026)Taghavi, Langari, and Pandey]{taghavi2026sparsepr}
Pardis Taghavi, Reza Langari, and Gaurav Pandey.
\newblock Partition the support, reconstruct the residual: Training-free sparse attention for video generation and world models, 2026.
\newblock URL \url{https://arxiv.org/abs/2608.18484}.

\bibitem[Tan et~al.(2026)Tan, Wang, Qiao, Zhang, Bai, Gong, Jin, and Li]{tan2026adacluster}
Haoyue Tan, Shengnan Wang, Yulin Qiao, Juncheng Zhang, Youhui Bai, Ping Gong, Zewen Jin, and Cheng Li.
\newblock {AdaCluster}: Adaptive query-key clustering for sparse attention in video generation.
\newblock In \emph{Proceedings of the IEEE/CVF Conference on Computer Vision and Pattern Recognition}, pp.\  43249--43259, 2026.

\bibitem[{Wan Team}(2025)]{wanteam2025wan}
{Wan Team}.
\newblock Wan: Open and advanced large-scale video generative models, 2025.
\newblock URL \url{https://arxiv.org/abs/2503.20314}.

\bibitem[Wang et~al.(2004)Wang, Bovik, Sheikh, and Simoncelli]{wang2004ssim}
Zhou Wang, Alan~C. Bovik, Hamid~R. Sheikh, and Eero~P. Simoncelli.
\newblock Image quality assessment: From error visibility to structural similarity.
\newblock \emph{IEEE Transactions on Image Processing}, 13\penalty0 (4):\penalty0 600--612, 2004.

\bibitem[Xi et~al.(2025)Xi, Yang, Zhao, Xu, Li, Li, Lin, Cai, Zhang, Li, Chen, Stoica, Keutzer, and Han]{xi2025sparsevideogen}
Haocheng Xi, Shuo Yang, Yilong Zhao, Chenfeng Xu, Muyang Li, Xiuyu Li, Yujun Lin, Han Cai, Jintao Zhang, Dacheng Li, Jianfei Chen, Ion Stoica, Kurt Keutzer, and Song Han.
\newblock {Sparse Video-Gen}: Accelerating video diffusion transformers with spatial-temporal sparsity.
\newblock In \emph{Proceedings of the 42nd International Conference on Machine Learning}, volume 267 of \emph{Proceedings of Machine Learning Research}, pp.\  68208--68224, 2025.

\bibitem[Xia et~al.(2025)Xia, Ling, Fu, Wang, Li, Xiao, and Cui]{xia2025training}
Yifei Xia, Suhan Ling, Fangcheng Fu, Yujie Wang, Huixia Li, Xuefeng Xiao, and Bin Cui.
\newblock Training-free and adaptive sparse attention for efficient long video generation.
\newblock In \emph{2025 IEEE/CVF International Conference on Computer Vision (ICCV)}, pp.\  15982--15993. IEEE, 2025.

\bibitem[Xu et~al.(2025)Xu, Xiao, Huang, Guo, and Han]{xu2025xattention}
Ruyi Xu, Guangxuan Xiao, Haofeng Huang, Junxian Guo, and Song Han.
\newblock {XA}ttention: Block sparse attention with antidiagonal scoring.
\newblock In \emph{Proceedings of the 42nd International Conference on Machine Learning}, volume 267 of \emph{Proceedings of Machine Learning Research}, pp.\  69819--69831. PMLR, 2025.
\newblock URL \url{https://proceedings.mlr.press/v267/xu25ag.html}.

\bibitem[Yang et~al.(2025{\natexlab{a}})Yang, Xi, Zhao, Li, Zhang, Cai, Lin, Li, Xu, Peng, Chen, Han, Keutzer, and Stoica]{yang2025svg2}
Shuo Yang, Haocheng Xi, Yilong Zhao, Muyang Li, Jintao Zhang, Han Cai, Yujun Lin, Xiuyu Li, Chenfeng Xu, Kelly Peng, Jianfei Chen, Song Han, Kurt Keutzer, and Ion Stoica.
\newblock {Sparse VideoGen2}: Accelerate video generation with sparse attention via semantic-aware permutation.
\newblock In \emph{Advances in Neural Information Processing Systems}, volume~38, pp.\  96965--96991, 2025{\natexlab{a}}.
\newblock \doi{10.52202/085713-3243}.

\bibitem[Yang et~al.(2025{\natexlab{b}})Yang, Teng, Zheng, Ding, Huang, Xu, Yang, Hong, Zhang, Feng, et~al.]{yang2025cogvideox}
Zhuoyi Yang, Jiayan Teng, Wendi Zheng, Ming Ding, Shiyu Huang, Jiazheng Xu, Yuanming Yang, Wenyi Hong, Xiaohan Zhang, Guanyu Feng, et~al.
\newblock {CogVideoX}: Text-to-video diffusion models with an expert transformer.
\newblock In \emph{International Conference on Learning Representations}, volume 2025, pp.\  83048--83077, 2025{\natexlab{b}}.

\bibitem[Zhang et~al.(2025)Zhang, Xiang, Huang, Wei, Xi, Zhu, and Chen]{zhang2025spargeattention}
Jintao Zhang, Chendong Xiang, Haofeng Huang, Jia Wei, Haocheng Xi, Jun Zhu, and Jianfei Chen.
\newblock {SpargeAttention}: Accurate and training-free sparse attention accelerating any model inference.
\newblock In \emph{Proceedings of the 42nd International Conference on Machine Learning}, volume 267 of \emph{Proceedings of Machine Learning Research}, pp.\  76397--76413, 2025.

\bibitem[Zhang et~al.(2018)Zhang, Isola, Efros, Shechtman, and Wang]{zhang2018lpips}
Richard Zhang, Phillip Isola, Alexei~A. Efros, Eli Shechtman, and Oliver Wang.
\newblock The unreasonable effectiveness of deep features as a perceptual metric.
\newblock In \emph{IEEE/CVF Conference on Computer Vision and Pattern Recognition (CVPR)}, pp.\  586--595, 2018.

\bibitem[Zhou et~al.(2026)Zhou, Mang, Yang, Xi, Zhang, Mao, Gonzalez, Keutzer, Stoica, and Cheung]{zhou2026svgear}
Xuanyi Zhou, Qiuyang Mang, Shuo Yang, Haocheng Xi, Jintao Zhang, Huanzhi Mao, Joseph~E. Gonzalez, Kurt Keutzer, Ion Stoica, and Alvin Cheung.
\newblock {SVG-EAR}: Parameter-free linear compensation for sparse video generation via error-aware routing.
\newblock \emph{arXiv preprint arXiv:2603.08982}, 2026.
\newblock URL \url{https://arxiv.org/abs/2603.08982}.
\newblock Accepted to ECCV 2026.

\end{thebibliography}

\clearpage
\appendix

\section{Supplementary Material}
\label{app:supplementary}

\subsection{Prompts}
\label{app:prompts}
Table~\ref{tab:prompts} lists the 22 VBench prompts, two from each of its eleven
per-dimension prompt lists, used verbatim for Table~\ref{tab:main},
Figure~\ref{fig:frontier} and Figure~\ref{fig:frames-quality}(a). The
importance-sampling generation times in Figure~\ref{fig:frontier} are measured on
the first five (d22-01 to d22-05).

\begin{table}[h]
  \centering
  \caption{\textbf{The 22 VBench prompts.}}
  \label{tab:prompts}
  \vspace{4pt}
  \small
  \begin{tabular}{@{}llp{0.62\linewidth}@{}}
    \toprule
    ID & VBench dimension & Prompt \\
    \midrule
    d22-01 & appearance style & a shark is swimming in the ocean by Hokusai, in the style of Ukiyo \\
    d22-02 & appearance style & A couple in formal evening wear going home get caught in a heavy downpour with umbrellas, in cyberpunk style \\
    d22-03 & color & a red car \\
    d22-04 & color & a blue bird \\
    d22-05 & human action & A person is riding or walking with horse \\
    d22-06 & human action & A person is catching or throwing frisbee \\
    d22-07 & multiple objects & a cat and a dog \\
    d22-08 & multiple objects & an airplane and a train \\
    d22-09 & object class & a train \\
    d22-10 & object class & a motorcycle \\
    d22-11 & overall consistency & Few big purple plums rotating on the turntable. water drops appear on the skin during rotation. isolated on the white background. close-up. macro. \\
    d22-12 & overall consistency & Balloon full of water exploding in extreme slow motion. \\
    d22-13 & scene & waterfall \\
    d22-14 & scene & highway \\
    d22-15 & spatial relationship & a skateboard on the top of a surfboard, front view \\
    d22-16 & spatial relationship & a traffic light on the left of a fire hydrant, front view \\
    d22-17 & subject consistency & a horse running to join a herd of its kind \\
    d22-18 & subject consistency & a bear catching a salmon in its powerful jaws \\
    d22-19 & temporal flickering & In a still frame, a tranquil pond was fringed by weeping cherry trees, their blossoms drifting lazily onto the glassy surface \\
    d22-20 & temporal flickering & A tranquil tableau of a tranquil lakeside cabin nestled among tall pines, its reflection mirrored perfectly in the calm water \\
    d22-21 & temporal style & An astronaut flying in space, racking focus \\
    d22-22 & temporal style & A panda drinking coffee in a cafe in Paris, zoom out \\
    \bottomrule
  \end{tabular}
\end{table}

Figure~\ref{fig:frames-quality}(b), the 81-frame check that sets the \method{}
budgets in it, and Figure~\ref{fig:floor-ablation} use a single prompt,
\texttt{svg2\_ex1}, the first example prompt of SVG2's released Wan2.1
script~\citep{yang2025svg2}:
\begin{quote}
\small warm colors dominate the room, with a focus on the tabby cat sitting contently in the center. the scene captures the fluffy orange tabby cat wearing a tiny virtual reality headset. the setting is a cozy living room, adorned with soft, warm lighting and a modern aesthetic. a plush sofa is visible in the background, along with a few lush potted plants, adding a touch of greenery. the cat's tail flicks curiously, as if engaging with an unseen virtual environment. its paws swipe at the air, indicating a playful and inquisitive nature, as it delves into the digital realm. the atmosphere is both whimsical and futuristic, highlighting the blend of analog and digital experiences.
\end{quote}

\end{document}